\pdfoutput=1
\documentclass[runningheads]{llncs}
\usepackage[T1]{fontenc}
\usepackage{graphicx}
\usepackage{booktabs}
\usepackage{multirow}
\usepackage{amsmath,amssymb}
\usepackage{enumitem}
\usepackage{xcolor}
\usepackage{algorithm,algpseudocode}
\usepackage{tcolorbox}
\tcbuselibrary{skins,breakable}
\usepackage{tikz}
\usepackage{pgfplots}
\pgfplotsset{compat=1.18}
\usetikzlibrary{calc,positioning,arrows.meta,shapes.geometric}
\usepackage{hyperref}

\definecolor{StageBg}{HTML}{F5F5F5}
\definecolor{StageStroke}{HTML}{B0B0B0}
\definecolor{BottleneckColor}{HTML}{D32F2F}
\definecolor{BottleneckBg}{HTML}{FFEBEE}
\definecolor{QueryColor}{HTML}{1F77B4}
\definecolor{StageNeutral}{HTML}{F2F2F2}
\definecolor{StageLabel}{HTML}{555555}

\begin{document}

\title{GRIP: Grounded Reasoning via\\ Information-Restricted Premises}
\titlerunning{GRIP: Grounded Reasoning via Information-Restricted Premises}

\author{Lirui Teng}
\authorrunning{L. Teng}
\institute{University of Waterloo, Waterloo ON, Canada\quad
\email{lteng@uwaterloo.ca}}

\maketitle

\sloppy

\begin{abstract}
High-capacity encoders in retrieval-augmented generation (RAG) can let
the query dominate the latent state, leaving retrieved evidence
functionally irrelevant. We call this failure mode query
dominance. To address it, we introduce \textbf{GRIP} (Grounded
Reasoning via Information-Restricted Premises), which imposes
capacity asymmetry: the decoder keeps full-dimensional access to
the query, while retrieved evidence passes through a severe stochastic
bottleneck. This forces the evidence channel to encode only the
residual information unavailable from the query. Across five reasoning
benchmarks, GRIP outperforms strong iterative baselines, cuts a
query--latent mutual-information diagnostic by roughly 30$\times$
(14.8 $\to$ 0.47 bits), and reduces hallucination by 73\%.
Residual-alignment analysis further shows that the bottleneck output
occupies subspaces less aligned with the query than baseline
representations.

\end{abstract}

\section{Introduction}
\label{sec:introduction}%

Retrieval-augmented generation (RAG) is intended to condition language
models on external evidence, approximating $P(Y \mid Q, E)$. In practice,
LLMs often under-use retrieved text and fall back on parametric
knowledge, even when it conflicts with the evidence~\cite{mallen2023trust}.
The failure is representational: $Q$ enters the decoder through a
high-capacity path while evidence shares the same latent space, so
optimization---already finding a low-loss solution under
$P(Y \mid Q)$---treats evidence as a marginal correction. We call the
resulting regime \emph{query dominance}.

Existing methods intervene at decoding or supervision time but leave the
latent geometry of query--evidence fusion largely unchanged. Self-RAG
adds reflection tokens for retrieval critique~\cite{asai2024selfrag};
context-aware decoding reweights token probabilities toward retrieved
content~\cite{shi2023trusting}; RAFT-style training teaches models to
ignore distractors~\cite{yoran2024making}. High-dimensional
representations can therefore still allocate most capacity to
query-aligned features and parametric
shortcuts~\cite{ethayarajh2019contextual,geirhos2020shortcut}, leaving
retrieval under-utilisation as a capacity-allocation problem rather than
only a content-selection one.

\begin{figure}[t]
\centering
\includegraphics[width=\textwidth]{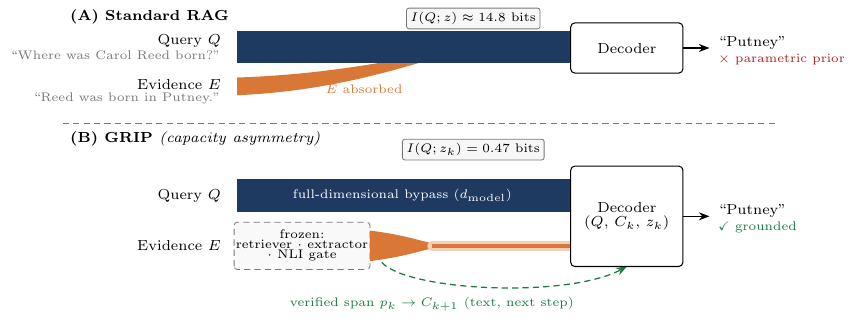}
\caption{Information flow in (A)~standard RAG versus (B)~GRIP. Ribbon
width is proportional to channel capacity. (A)~Evidence enters the
same full-dimensional latent as the query and is absorbed by the
query-aligned flow, leaving answers driven by the parametric prior.
(B)~The query retains a full-dimensional bypass while
evidence---produced by a frozen retrieve--extract--verify pipeline---is
squeezed through a stochastic $d_z{=}4$ bottleneck (${\approx}$2--4
bits/step); the entailment-verified span additionally persists as text
in $C_{k+1}$ (Sect.~\ref{sec:forward-pass}). Boxed values give
measured query--latent mutual information
(Table~\ref{tab:mechanism-diagnostics}).}
\label{fig:intro-pipeline}
\end{figure}

We address this gap through deliberate \emph{capacity asymmetry} rather
than richer fusion. \textbf{GRIP} (Grounded Reasoning via
Information-Restricted Premises; Fig.~\ref{fig:intro-pipeline}) routes
$Q$ through a full-dimensional bypass while forcing retrieved evidence
through an aggressively low-dimensional, stochastic bottleneck
($d_z \approx 4$). Because the decoder retains high-bandwidth access to
$Q$, query-correlated bits in the bottleneck are redundant under a tight
capacity budget; gradients pressure the channel to transmit only the
\emph{information residual}---the signal the query does not already
provide. Per query, GRIP operates on a cumulative total of roughly 25
entailment-verified evidence tokens across its two reasoning steps
(averaging ${\sim}11$ tokens per step), compared to the baseline's
$\sim$4{,}000-token raw retrieved context.

\paragraph{Contributions.}
First, we formalise query dominance as a failure of conditional
independence and introduce \emph{Query--Latent (QL) Dependence}, the
mutual information $I(Q; z_k)$ between the query and the evidence
representation, as a model-agnostic diagnostic: elevated QL dependence
indicates that $z_k$ has collapsed into a compressed copy of the query
and coincides with elevated hallucination. Second, we introduce GRIP, which enforces low-capacity,
noise-regularised evidence representations that we argue are consistent
with an information-residual mechanism. Third, GRIP outperforms strong
baselines on HotpotQA, StrategyQA, 2Wiki, ProofWriter, and SQuAD~2.0,
reducing QL dependence by roughly $20$--$37\times$, suppressing
hallucination by 73\%, and producing residuals more orthogonal to the
query than unconstrained baselines.

\section{Query Dominance in Latent States}
\label{sec:query-dominant}

Although RAG is intended to approximate $P(Y \mid Q,E)$, language
models in practice often behave closer to $P(Y \mid Q)$: they generate
from parametric knowledge while showing limited sensitivity to the
retrieved
context~\cite{bi2025confiqa,joren2025sufficient,mallen2023trust,shi2023trusting,yoran2024making},
a phenomenon that recent analyses show worsens with model capacity. We
refer to the representational form of this failure as \emph{query
dominance}: the latent state used by the decoder remains highly
predictable from the query and only weakly responsive to evidence
variation: retrieval is present in the pipeline but functionally
marginal in generation.

\subsection{The Parametric Prior Pathology}
\label{sec:bypass-intuition}

This failure is encouraged by the geometry of pretrained transformer
representations: contextual embeddings are anisotropic, with a few
high-variance directions carrying broad semantic and frequency
information~\cite{ethayarajh2019contextual}. Query features tend to occupy
these dominant directions, so retrieved evidence---even when relevant---is
treated as a perturbation to an already strong query-conditioned
trajectory, a form of shortcut learning in which a dominant signal
suppresses genuinely joint representations~\cite{geirhos2020shortcut}.

\subsection{Diagnosing Query Dominance}
\label{sec:leakage}

We model an evidence-conditioned system as
\begin{equation}
\hat{Y} = G(z,Q), \qquad z=\phi(Q,E),
\label{eq:model-env}
\end{equation}
where $z$ is the internal evidence representation passed to the decoder.
Let $F(Q,E)=G(\phi(Q,E),Q)$ denote the end-to-end output map. We assume
that, for fixed $Q$, the effect of evidence on the output is mediated
through $z$; that $G(\cdot,Q)$ is locally Lipschitz under the chosen output
discrepancy metric $d$; and that representations are bounded. Under
these assumptions, when contrastive evidence produces little separation
in latent space, the decoder cannot produce large output separation
either: query dominance can arise from collapse or redundancy in
$\phi(Q,E)$ itself.

For a fixed query $q$, let $E_q^+$ and $E_q^-$ denote contrastive evidence
distributions that support different task-level answers. We define
\emph{contrastive evidence sensitivity} as
\begin{equation}
S_E(q)
=
\mathbb{E}_{e^+\sim E_q^+,\,e^-\sim E_q^-}
\left[
d\!\left(F(q,e^+),F(q,e^-)\right)
\right].
\label{eq:evidence-sensitivity}
\end{equation}
A model is behaviourally query-dominant at $q$ when $S_E(q)$ is small while
the model remains sensitive to changes in the query. To rule out trivial
constant-output collapse, we define query-swap sensitivity
$S_Q(q) = \mathbb{E}_{q'\sim \mathcal{Q},\,e\sim E_q^+}
[\, d(F(q,e),F(q',e)) \,]$
and say that query dominance occurs when $S_E(q)\le \tau_E$ and
$S_Q(q)\ge \tau_Q$ for thresholds $\tau_E \ll \tau_Q$.

At the representation level, we measure \emph{Query--Latent (QL)
Dependence}:
\begin{equation}
D_{\mathrm{QL}}
=
I(Q;z_k).
\label{eq:ql-dependence}
\end{equation}
High $D_{\mathrm{QL}}$ indicates that the evidence-channel state $z_k$ is
strongly predictable from the query, suggesting that it carries
query-aligned information rather than evidence-specific content. In a
well-conditioned retrieval system, $z_k$ should instead encode conditional
innovation: information supplied by evidence that is not already available
from $Q$.
We estimate $I(Q;z_k)$ using the Contrastive Log-ratio Upper Bound (CLUB)
estimator~\cite{cheng2020club}. Because neural MI estimators exhibit
dimension-dependent bias~\cite{czyz2023beyond} and can violate basic
self-consistency properties~\cite{song2020understanding}, we rely on
relative comparisons across conditions rather than absolute values. A
complementary shuffle control---breaking query--evidence correspondence
and verifying that the estimate collapses toward zero---can further
validate the estimator. QL dependence is a necessary but not
sufficient diagnostic---randomization, collapse, or decoder null-space
effects can also reduce $I(Q;z_k)$---so we pair it with behavioural
randomization tests and residual-alignment measurements in
Section~\ref{sec:experiments}.

\section{Design Principle: Capacity-Asymmetric Evidence}
\label{sec:capacity-evidence}

Given the query-dominant regime described above, the central design question
is not only which evidence to retrieve, but how much representational
capacity the evidence pathway should receive. GRIP adopts a
capacity-asymmetric principle: the query and reasoning context retain
full-dimensional access to the decoder, while evidence is routed through a
deliberately restricted channel. The goal is not to remove the query signal,
but to prevent the evidence representation from cheaply duplicating
information already available through the query path. GRIP thus differs
from evidence-compression methods such as xRAG, COCOM, PISCO, and gist
tokens, which compress context to maximise retention and efficiency,
whereas GRIP restricts evidence capacity to counteract query dominance.

\paragraph{Information-bottleneck approaches to RAG.} Zhu et
al.~\cite{zhu2024ib} filter retrieval noise by maximising mutual
information between a compressed representation and the output while
minimising it with the passage---a largely deterministic noise filter.
Swin-VIB~\cite{wang2025swinvib} integrates variational IB models that
adaptively regulate evidence compression to guide an LLM under knowledge
conflicts. GRIP differs in three respects: (i) deliberate capacity
\emph{asymmetry} (full-dimensional query access versus a severely
restricted evidence channel) rather than an adaptive conflict adapter;
(ii) a fixed, severe stochastic bottleneck ($d_z{=}4$, additive
Gaussian noise, ${\approx}2$--$4$ bits per step) as a first-class design
principle rather than a learned compression ratio; and (iii) a
motivation of counteracting query dominance rather than arbitrating
knowledge conflicts.

\subsection{Capacity-Limited Evidence Representations}
\label{sec:bottleneck-principle}

Abstractly, GRIP maps an extracted, verified premise $p_k$ to a
low-dimensional noisy state
\begin{equation}
z_k = B_\theta(p_k)+\varepsilon_k,
\qquad
\varepsilon_k\sim\mathcal{N}(0,\sigma^2 I_{d_z}),
\qquad
d_z\ll d_{\mathrm{model}},
\label{eq:abstract-bottleneck}
\end{equation}
where $B_\theta$ denotes the evidence compressor. Under the standard
Gaussian channel approximation, $z_k$ cannot transmit unrestricted premise
information; Section~\ref{sec:bottleneck} gives the explicit capacity bound
at the implementation level. Low dimensionality limits transmittable
features, premise-level compression reduces passage verbosity, and
additive noise discourages deterministic copying of brittle
correlations---together making it inefficient for $z_k$ to serve as a
second query representation.

\subsection{Residualization Pressure}
\label{sec:residualization-pressure}

Under a tight evidence capacity budget, query-correlated information in
$z_k$ has an opportunity cost: capacity spent on query-predictable
features is inefficient unless those features also help predict $Y$
given $Q$---exactly the trade-off captured by the Conditional
Information Bottleneck objective:
\begin{equation}
\mathcal{L}_{\mathrm{CIB}}
=
-I(Z_k;Y\mid Q)
+
\beta I(Z_k;Q).
\label{eq:cib-pressure}
\end{equation}
GRIP does not optimize Eq.~\eqref{eq:cib-pressure} explicitly; the
capacity asymmetry creates a similar pressure---preserve evidence
information predictive of $Y$ while discouraging redundant query
information---which we treat as a mechanism-level interpretation, not a
formal equivalence.

\subsection{Expected Diagnostic Effects}
\label{sec:testable-consequences}

The capacity-asymmetry hypothesis yields three empirical predictions.
First, query--latent dependence should decrease:
$I(Q;z_k)_{\mathrm{GRIP}} \ll I(Q;z_k)_{\mathrm{RAG}}$.
Second, if the decoder genuinely relies on the restricted evidence channel,
randomizing $z_k$ should cause a larger performance drop than the
analogous intervention on baselines:
$\Delta_{\mathrm{rand}}^{\mathrm{GRIP}} > \Delta_{\mathrm{rand}}^{\mathrm{baseline}}$,
where $\Delta_{\mathrm{rand}} = \mathrm{Acc}(z_k) - \mathrm{Acc}(\tilde{z}_k)$
and $\tilde{z}_k$ is a randomized bottleneck state. Third,
residual-alignment measurements should show that $z_k$ occupies subspaces
less aligned with query-dominant directions than baseline representations.

\begin{figure}[t]
\centering
\includegraphics[width=0.49\textwidth]{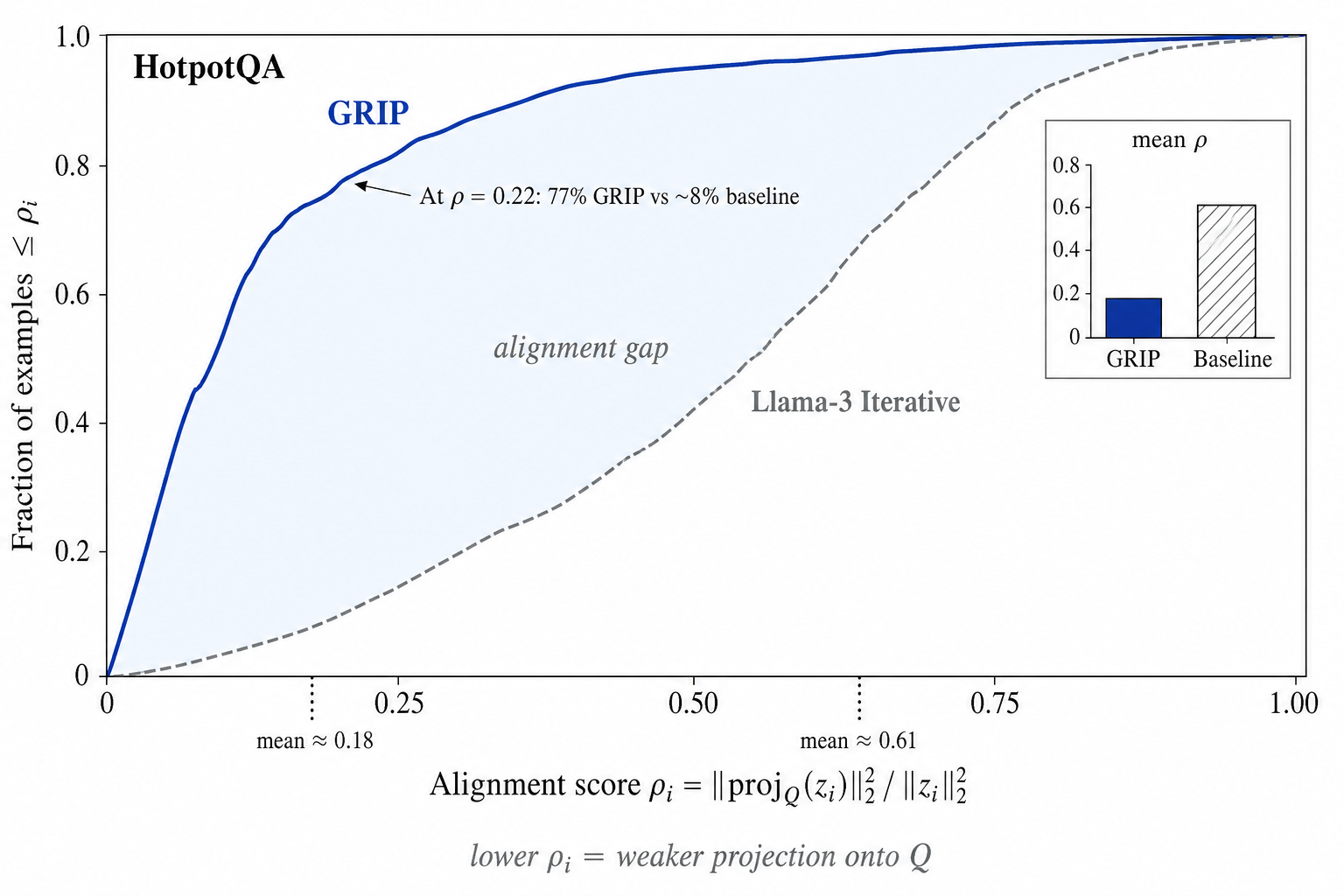}
\caption{\textbf{Empirical CDFs of evidence--query alignment on
HotpotQA} (lower $\rho$ $\Rightarrow$ weaker query alignment). The GRIP
distribution (solid, deep blue) dominates the Llama-3 Iterative baseline
(dashed, gray) at every threshold: at $\rho = 0.22$, 77\% of GRIP samples
fall below this threshold versus $\sim$8\% of baseline samples, so the
reduction is systematic rather than outlier-driven. Inset: mean $\rho$
per method ($\rho$ column of Table~\ref{tab:mechanism-diagnostics}).}
\label{fig:alignment-ecdf}
\end{figure}

Section~\ref{sec:experiments} tests all three predictions: QL dependence
and $\Delta_{\mathrm{rand}}$ across the five benchmarks, with the
architecture-matched Llama-3 Iterative control on HotpotQA, and $\rho$
in Fig.~\ref{fig:alignment-ecdf} and
Table~\ref{tab:mechanism-diagnostics}. Convergence of the three
diagnostics, together with the ablation pattern of
Section~\ref{sec:ablation}, supports the capacity-asymmetry account.

\section{Architecture}
\label{sec:architecture}

\begin{figure}[t]
\centering
\includegraphics[width=\textwidth]{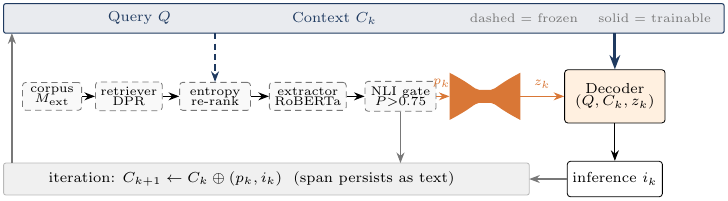}
\caption{GRIP implementation pipeline: at each step, candidate passages
are retrieved and entropy-ranked, reduced to a predictive span, filtered
by an NLI verifier, compressed to $z_k$, and passed to the decoder
alongside the full-dimensional query and context.}
\label{fig:grip-architecture}
\end{figure}

The capacity-asymmetric principle of Section~\ref{sec:capacity-evidence}
is realised as a four-stage pipeline (Fig.~\ref{fig:grip-architecture}):
iterative retrieval, predictive span extraction, stochastic compression,
and asymmetric decoding. The retriever, extractor, and verifier are
frozen; gradients flow only through the bottleneck and decoder.

\subsection{Retrieval via Entropy-Guided Re-ranking}
\label{sec:retrieval}

A dense passage retriever~\cite{karpukhin2020dpr} returns top-$m$
candidates, which are re-ranked by the conditional entropy of the
next-step prediction:
\begin{equation}
s(r \mid C_k, Q) \;=\; - H_\Theta\!\bigl(i_k \mid r, C_k, Q\bigr).
\label{eq:retrieval-score}
\end{equation}
Passages making the next step more predictable score higher; entropy is
computed under teacher-forced decoding through the candidate. A
curriculum defers the entropy criterion until the decoder is calibrated
(Section~\ref{sec:training-objective}).

\subsection{Predictive Span Extraction}
\label{sec:extraction}

The selected passage $r^{*(k)}$ is reduced to a minimal predictive span
before it reaches the bottleneck. A frozen RoBERTa-based extractor
$\Theta_{\mathrm{ext}}$ produces $p_k \subset r^{*(k)}$ by KL-matching
the next-step distribution conditioned on the span versus the full
passage, with a length-sparsity penalty (full objective in the
supplement). A frozen DeBERTa-v3-large NLI verifier admits only spans
satisfying $P_{\mathrm{NLI}}(\mathrm{entailment} \mid r^{*(k)}, p_k) >
0.75$. To prevent premise representations from becoming query-dependent,
query tokens are masked in the extractor's cross-attention, isolating
$\Theta_{\mathrm{ext}}$'s gradients from query embeddings.

\subsection{Stochastic Bottleneck}
\label{sec:bottleneck}

The verified span is then mapped to a low-dimensional latent through a
noisy projection:
\begin{equation}
z_k = W_2 \, \sigma\!\bigl(W_1 \cdot \mathrm{pool}(p_k)\bigr) + \varepsilon_k,
\quad \varepsilon_k \sim \mathcal{N}(0, \sigma^2 I_{d_z}),
\label{eq:bottleneck}
\end{equation}
with $d_z = 4$, $\sigma^2 = 1.0$, mean pooling, and ReLU activation. For
a $d_z$-dimensional channel with additive Gaussian noise, the Gaussian
channel capacity gives
\begin{equation}
I(\mathrm{pool}(p_k); z_k)
\;\le\;
\tfrac{d_z}{2} \log\!\bigl(1 + P/\sigma^2\bigr),
\label{eq:capacity-bound}
\end{equation}
where $P$ is the average power of the projected pre-noise vector. With
$P \approx 1$ after normalisation, the per-step budget is approximately
2--4 bits---the explicit form of the capacity asymmetry argued for in
Section~\ref{sec:capacity-evidence}.

\subsection{Asymmetric Decoding}
\label{sec:decoder}

The decoder conditions on $(Q, C_k, z_k)$: the query enters as a
high-bandwidth prefix, the context $C_k$ carries prior reasoning steps,
and $z_k$ is injected as a single special token with a learned
positional embedding. Because $Q$ and $C_k$ enter at full dimensionality
while $z_k$ is constrained to four dimensions with additive noise, the
query/context and bottleneck pathways differ in capacity by roughly
three orders of magnitude.

\subsection{Forward-Pass Procedure}
\label{sec:forward-pass}

Algorithm~\ref{alg:grip-forward} composes the four stages into a
step-wise inference loop; entropy-guided selection
(Eq.~\eqref{eq:retrieval-score}) is a non-differentiable inference-time
decision based on the decoder's state.

\algnewcommand{\LineComment}[1]{%
  \Statex \hskip\algorithmicindent\(\triangleright\)~\textit{#1}%
}

\begin{algorithm}[t]
\footnotesize
\caption{GRIP Step-Wise Inference}
\label{alg:grip-forward}
\begin{algorithmic}[1]
\Require Query $Q$; corpus $M_{\mathrm{ext}}$; frozen modules
         (retriever, $\Theta_{\mathrm{ext}}$, $\mathrm{NLI}$);
         trained modules ($\mathrm{Bottleneck}$, $\mathrm{Decode}$);
         max steps $K$; entailment threshold $\tau$
\Ensure  Final answer $\hat{a}$
\State $C_1 \gets \emptyset$
   \Comment{empty reasoning context}
\For{$k = 1, \dots, K$}
    \State $R_m^{(k)} \gets \mathrm{DenseRetrieve}(Q, C_k)$
    \State $r^{*(k)} \gets \arg\min_{r \in R_m^{(k)}} H_\Theta(i_k \mid r, C_k, Q)$

    \State $p_k \gets \Theta_{\mathrm{ext}}(r^{*(k)}, C_k)$
    \If{$\mathrm{NLI}(p_k, r^{*(k)}) < \tau$}
        \State \textbf{continue}
            \Comment{discard step $k$; advance with context unchanged}
    \EndIf

    \State $z_k \gets \mathrm{Bottleneck}(p_k) + \varepsilon_k,
           \quad \varepsilon_k \sim \mathcal{N}(0, \sigma^2 I_{d_z})$

    \State $i_k \gets \mathrm{Decode}(Q, C_k, z_k)$
    \State $C_{k+1} \gets C_k \oplus (p_k, i_k)$
    \If{$i_k$ emits \textsc{[answer]}}
        \State \Return $i_k$
            \Comment{early termination}
    \EndIf
\EndFor
\State \Return $i_K$
   \Comment{fallthrough: no \textsc{[answer]} within $K$ steps}
\end{algorithmic}
\end{algorithm}

The context update $C_{k+1} \leftarrow C_k \oplus (p_k, i_k)$ retains
the verified span as text, so the decoder keeps a full-dimensional
semantic pathway alongside the bottleneck. This is deliberate rather
than a leak: ablating the raw-span pathway costs $8.2$ accuracy points
(Section~\ref{sec:ablation}), while at step $k$ the decoder commits to
inference $i_k$ before $p_k$ becomes contextually available, and
randomising $z_k$ still costs $35.3$ points with a $71.6{:}1$ ratio of
correct-to-wrong versus wrong-to-correct transitions
(Section~\ref{sec:diagnostics}). The bottleneck thus regulates the
evidence signal governing each inference step; the persistent span text
carries sentence-level semantics but cannot substitute for the
bottleneck state. What persists is the minimal entailment-verified
span, not the retrieved passage.

\subsection{Training Objective and Schedule}
\label{sec:training-objective}

The trainable parameters $\Theta_{\mathrm{gen}} = \{W_1, W_2,
\Theta_{\mathrm{dec}}\}$ maximise the likelihood of the target
reasoning steps,
\begin{equation}
\mathcal{L}_{\text{task}}
\;=\;
-\sum_{k=1}^{K} \log P_{\Theta_{\mathrm{gen}}}(i_k \mid z_k, C_k, Q),
\label{eq:joint-loss}
\end{equation}
under a two-phase curriculum. In Phase~1 (epochs 1--5) the entropy
re-ranker is bypassed and top passages are selected by dense retrieval
alone, allowing the bottleneck and decoder to stabilise before the
entropy signal becomes load-bearing. In Phase~2 (epochs 6--20),
entropy-guided selection (Eq.~\eqref{eq:retrieval-score}) is enabled
while the retriever, extractor, and NLI verifier remain frozen
throughout. Because $Q$ reaches the decoder at full dimensionality
through the bypass, query-redundant features in $z_k$ do not reduce the
conditional likelihood and are pruned by gradient
descent~\cite{achille2018emergence}.

\section{Experiments}
\label{sec:experiments}

We evaluate GRIP on five reasoning benchmarks to test whether
capacity-asymmetric evidence processing improves task performance and
evidence use.

\subsection{Experimental Configuration}
\label{sec:setup}

\paragraph{Baselines.}
We compare GRIP against three baselines spanning the matched-control
and prior-art axes. \textbf{Standard RAG} uses DPR
retrieval~\cite{karpukhin2020dpr} with passage concatenation and
Llama-3-8B decoding~\cite{lewis2020rag}. \textbf{Self-Ask}~\cite{press2023selfask}
uses iterative sub-question prompting without modifying the evidence
pathway.
\textbf{Llama-3-8B Iterative} is the architecture-matched control: it
follows GRIP's two-step reasoning schedule ($K{=}2$, same retriever,
entropy re-ranking, span extraction, and NLI gate) but injects the
verified premise $p_k$ into the decoder as ordinary full-dimensional
text rather than through the stochastic bottleneck, isolating the
contribution of capacity asymmetry from that of iterative reasoning.
Decoding hyperparameters match GRIP.

All systems share the same frozen DPR-Wiki index, tokenization, and
compute budget. GRIP performs $K=2$ reasoning steps with $m=10$
retrieved passages per step. The shared component is the retrieval
substrate (retriever and index); the evidence reaching each decoder
still differs after re-ranking, extraction, and NLI filtering, so the
comparison isolates how each method processes evidence.
Descriptive statistics of the retrieval--verification pipeline are
reported in the supplement.

\paragraph{Datasets.}
We evaluate on five benchmarks covering different reasoning regimes:
\textbf{HotpotQA}~\cite{yang2018hotpotqa} for distractor multi-hop QA,
\textbf{StrategyQA}~\cite{geva2021strategyqa} for implicit multi-step
reasoning, \textbf{2WikiMultihopQA}~\cite{ho2020constructing} for
explicit two-hop reasoning,
\textbf{ProofWriter}~\cite{tafjord2021proofwriter} for symbolic
Horn-clause deduction, and
\textbf{SQuAD~2.0}~\cite{rajpurkar2018squad} for single-hop extractive
QA. Primary metrics are exact match (EM) or task accuracy, with F1
where applicable (Table~\ref{tab:main-results}). Hallucination
is the percentage of generated claims not entailed by retrieved
evidence. The in-pipeline DeBERTa-v3 verifier that scores entailment is
also the training-time selection signal; an independent verifier
provides an evaluation check on this circularity
(Section~\ref{sec:main-results}). Atomicity of extracted premises is
reported in the supplement.

\paragraph{Optimization.}
Trainable components use AdamW~\cite{loshchilov2019decoupled}
(lr $10^{-4}$, weight decay $0.01$, 1{,}000-step warmup, cosine decay,
gradient clipping 1.0, global batch 128), with nucleus sampling
($p{=}0.9$, $T{=}0.7$), trained for 20 epochs on $4\times$A100 80GB
GPUs. Reported results are averaged
over three random seeds; on HotpotQA the standard deviation over seeds
is $\pm 0.45$ EM, $\pm 0.38$ F1, $\pm 0.72$ hallucination, and
$\pm 0.04$ bits QL dependence.

\subsection{Main Results}
\label{sec:main-results}

\begin{table}[htbp]
\centering
\footnotesize
\renewcommand{\arraystretch}{0.90}
\setlength{\tabcolsep}{8pt}
\caption{Task performance across five reasoning benchmarks. EM/F1
follow standard conventions for HotpotQA, 2Wiki, and SQuAD~2.0;
StrategyQA reports accuracy only (yes/no); ProofWriter reports proof
accuracy. Em dashes denote metrics not applicable (accuracy-only
tasks).}
\label{tab:main-results}
\begin{tabular}{llccc}
\toprule
\textbf{Dataset} & \textbf{Model}
& \textbf{EM/Acc} & \textbf{F1}
& \textbf{Hall.~(\%)}~$\downarrow$ \\
\midrule
\multirow{4}{*}{HotpotQA}
  & Standard RAG       & 68.2 & 72.5 & 31.7 \\
  & Self-Ask           & 71.3 & 75.8 & 19.8 \\
  & Llama-3 Iterative  & 69.3 & 73.6 & 28.7 \\
  & \textbf{GRIP}      & \textbf{76.5} & \textbf{80.3} & \textbf{8.6} \\
\addlinespace[2pt]
\multirow{4}{*}{StrategyQA}
  & Standard RAG       & 65.2 & --   & 33.4 \\
  & Self-Ask           & 67.5 & --   & 32.1 \\
  & Llama-3 Iterative  & 69.3 & --   & 28.7 \\
  & \textbf{GRIP}      & \textbf{73.4} & --   & \textbf{10.1} \\
\addlinespace[2pt]
\multirow{4}{*}{2Wiki}
  & Standard RAG       & 62.8 & 68.4 & 31.2 \\
  & Self-Ask           & 66.3 & 71.5 & 19.7 \\
  & Llama-3 Iterative  & 64.8 & 69.8 & 28.4 \\
  & \textbf{GRIP}      & \textbf{71.2} & \textbf{76.1} & \textbf{9.8} \\
\addlinespace[2pt]
\multirow{4}{*}{ProofWriter}
  & Standard RAG       & 74.3 & --   & 26.1 \\
  & Self-Ask           & 78.5 & --   & 15.9 \\
  & Llama-3 Iterative  & 77.0 & --   & 23.7 \\
  & \textbf{GRIP}      & \textbf{85.6} & --   & \textbf{6.8} \\
\addlinespace[2pt]
\multirow{4}{*}{SQuAD~2.0}
  & Standard RAG       & 78.4 & 82.1 & 18.2 \\
  & Self-Ask           & 76.5 & 80.3 & 19.4 \\
  & Llama-3 Iterative  & 78.0 & 82.3 & 17.7 \\
  & \textbf{GRIP}      & \textbf{82.1} & \textbf{85.7} & \textbf{6.4} \\
\bottomrule
\end{tabular}
\end{table}

Table~\ref{tab:main-results} reports task performance across all five
datasets. GRIP improves over the strongest non-GRIP baseline on every
dataset, and outperforms the architecture-matched Llama-3 Iterative control on
all five benchmarks---by $+7.2$ EM on HotpotQA and $+4.1$ accuracy
points on StrategyQA---indicating that capacity asymmetry contributes
beyond the iterative reasoning schedule. Paired bootstrap comparisons
are significant at $p<0.01$ on HotpotQA ($+7.2$) and SQuAD~2.0
($+3.7$).
Self-Ask is competitive on the multi-hop settings but trails Standard
RAG on single-hop SQuAD~2.0 (76.5 vs.\ 78.4 EM), consistent with
decomposition overhead when explicit multi-hop decomposition is
unnecessary. Hallucination drops substantially in every
condition---from 31.7\% to 8.6\% on HotpotQA, from 31.2\% to 9.8\% on
2Wiki, and from 28.7\% (Llama-3 Iterative) to 8.6\% (GRIP) under the
matched control. This grounding result is robust to the choice of
verifier: rescoring with MiniCheck~\cite{tang2024minicheck} yields
89.0\% agreement with the in-pipeline verifier (Cohen's
$\kappa = 0.77$), and
the HotpotQA hallucination rate rises only from 8.6\% to 10.1\%,
remaining well below every baseline---the gains are thus not
verifier-specific, though this does not establish model-agnosticism.

\subsection{Ablation Studies}
\label{sec:ablation}

Table~\ref{tab:ablation-combined} reports component and capacity
ablations on HotpotQA. The bottleneck is the most load-bearing
component: removing it raises QL dependence from 0.47 to 14.20 bits
and reduces accuracy by 5.3 points. Removing extraction or NLI
verification degrades performance through different failure modes:
without extraction, verbose passage content enters the bottleneck and
hallucination rises; without NLI verification, unsupported spans are
admitted, raising hallucination while leaving QL dependence low.

\begin{table}[htbp]
\centering
\footnotesize
\renewcommand{\arraystretch}{0.92}
\setlength{\tabcolsep}{6pt}
\caption{Component and capacity ablations on HotpotQA. $\Delta$
reports accuracy change relative to full GRIP. Component ablations
isolate individual modules; capacity ablations vary the bottleneck
width $d_z$ at fixed noise $\sigma^2=1.0$. For ``No bottleneck'',
$I(Q;z_k)$ is measured on the pooled premise embedding that replaces
$z_k$ (no low-rank projection or noise).}
\label{tab:ablation-combined}
\begin{tabular}{lcccr}
\toprule
\textbf{Configuration}
& $I(Q;z_k)$~\textbf{(bits)}~$\downarrow$
& \textbf{Hall.~(\%)}~$\downarrow$
& \textbf{Acc.}
& $\boldsymbol{\Delta}$ \\
\midrule
\textbf{Full GRIP} ($d_z=4$)
& \textbf{0.47} & \textbf{8.6} & \textbf{76.5} & -- \\
\midrule
\multicolumn{5}{l}{\textit{Component ablations}} \\
\addlinespace[2pt]
No bottleneck            & 14.20 & 24.3 & 71.2 & $-5.3$ \\
No extraction            &  3.10 & 18.7 & 73.4 & $-3.1$ \\
No NLI verification      &  0.52 & 12.1 & 75.1 & $-1.4$ \\
\midrule
\multicolumn{5}{l}{\textit{Capacity ablations}} \\
\addlinespace[2pt]
$d_z=2$                  & 0.31 & 9.2  & 74.2 & $-2.3$ \\
$d_z=8$                  & 1.82 & 8.9  & 75.1 & $-1.4$ \\
$d_z=16$                 & 4.73 & 11.5 & 72.8 & $-3.7$ \\
\midrule
\multicolumn{5}{l}{\textit{Mechanism controls}} \\
\addlinespace[2pt]
Deterministic ($d_z=4$, $\sigma^2{=}0$) & 2.38  & 13.2 & 74.0 & $-2.5$ \\
Noise-only ($d{=}4096$, stochastic)     & 10.85 & 24.7 & 72.1 & $-4.4$ \\
\bottomrule
\end{tabular}
\end{table}

The capacity sweep is non-monotonic. A very narrow bottleneck
($d_z=2$) suppresses QL dependence most strongly but loses
task-relevant evidence; a wider one ($d_z=16$) restores capacity but
allows query-redundant information to re-enter. The best
configuration is therefore not the smallest channel, but the channel
that balances residual evidence transmission against query redundancy.

The mechanism controls in Table~\ref{tab:ablation-combined} follow the
deterministic-versus-stochastic pairing of information-bottleneck
designs in prior RAG work~\cite{wang2025swinvib,zhu2024ib}. Neither
dimensional restriction nor stochastic corruption alone is sufficient:
at fixed $d_z=4$, removing stochasticity increases QL dependence from
0.47 to 2.38 bits and hallucination by 4.6 points, while retaining
stochasticity at full dimension ($d=4096$) raises QL dependence to
10.85 bits and hallucination to 24.7\%. The combination of restricted
capacity and stochastic encoding is therefore critical to GRIP's
information-control behaviour.

Two bypass ablations complete the picture. Removing the query bypass
(the decoder receives $z_k$ and $C_k$ but no full-dimensional access to
$Q$) causes severe generation degradation of 23--33 accuracy points
across the five datasets even though QL dependence stays at 0.45 bits:
the 4D stochastic channel alone is too narrow to carry the semantic
burden of generation. Removing the raw-text bypass instead (the decoder
receives $Q$, the inference history, and $z_k$, but no persisted span
text) costs 8.2 points on HotpotQA/2Wiki and raises hallucination to
18.4\%. The architecture thus requires both restricted evidence flow
and high-capacity semantic access.

\subsection{Mechanism Diagnostics}
\label{sec:diagnostics}

Because aggregate accuracy does not reveal whether the bottleneck
changes evidence use, we report three diagnostics: QL dependence
$I(Q; z_k)$, estimated with the CLUB upper
bound~\cite{cheng2020club}; the randomization drop
$\Delta_{\mathrm{rand}} = \mathrm{Acc}(z_k) - \mathrm{Acc}(\tilde{z}_k)$,
where $\tilde{z}_k$ is sampled from the empirical marginal of
bottleneck states; and residual alignment
$\rho(z_k, \mathcal{Q}) = \|\mathrm{proj}_{\mathcal{Q}}(z_k)\|_2^2 / \|z_k\|_2^2$,
where $\mathcal{Q}$ is the top principal-component subspace of query
embeddings (90\% of variance, 10K-sample validation pool).
Table~\ref{tab:mechanism-diagnostics} consolidates all three
diagnostics across the five datasets for GRIP, the matched control,
and Standard RAG.

\begin{table}[htbp]
\centering
\footnotesize
\renewcommand{\arraystretch}{0.90}
\setlength{\tabcolsep}{9pt}
\caption{Mechanism diagnostics across datasets, as defined in
Section~\ref{sec:diagnostics}: lower $I(Q;z_k)$ $\Rightarrow$ less
query-redundant representation; higher $\Delta_{\mathrm{rand}}$
$\Rightarrow$ stronger decoder dependence on evidence; lower $\rho$
$\Rightarrow$ weaker geometric query alignment. Em dashes denote
diagnostics not run ($\Delta_{\mathrm{rand}}$ for Standard RAG).}
\label{tab:mechanism-diagnostics}
\begin{tabular}{llccc}
\toprule
\textbf{Dataset} & \textbf{Model}
& $I(Q;z_k)$~\textbf{(bits)}~$\downarrow$
& $\boldsymbol{\Delta_{\mathrm{rand}}}$~$\uparrow$
& $\boldsymbol{\rho}$~$\downarrow$ \\
\midrule
\multirow{3}{*}{HotpotQA}
  & Standard RAG       & 14.8 & --   & 0.72 \\
  & Llama-3 Iterative  & 11.2 & 7.5  & 0.61 \\
  & \textbf{GRIP}      & \textbf{0.47} & \textbf{35.3} & \textbf{0.18} \\
\addlinespace[2pt]
\multirow{3}{*}{StrategyQA}
  & Standard RAG       & 13.2 & --   & 0.68 \\
  & Llama-3 Iterative  & 10.1 & 4.2  & 0.58 \\
  & \textbf{GRIP}      & \textbf{0.52} & \textbf{30.5} & \textbf{0.19} \\
\addlinespace[2pt]
\multirow{3}{*}{2Wiki}
  & Standard RAG       & 15.1 & --   & 0.74 \\
  & Llama-3 Iterative  & 11.5 & 6.8  & 0.63 \\
  & \textbf{GRIP}      & \textbf{0.41} & \textbf{35.0} & \textbf{0.17} \\
\addlinespace[2pt]
\multirow{3}{*}{ProofWriter}
  & Standard RAG       & 11.8 & --   & 0.69 \\
  & Llama-3 Iterative  &  9.2 & 8.1  & 0.56 \\
  & \textbf{GRIP}      & \textbf{0.38} & \textbf{42.5} & \textbf{0.13} \\
\addlinespace[2pt]
\multirow{3}{*}{SQuAD~2.0}
  & Standard RAG       & 12.4 & --   & 0.78 \\
  & Llama-3 Iterative  & 10.4 & 5.5  & 0.65 \\
  & \textbf{GRIP}      & \textbf{0.61} & \textbf{22.5} & \textbf{0.24} \\
\bottomrule
\end{tabular}
\end{table}

GRIP reduces QL dependence by $20\times$--$37\times$ across all five
benchmarks, tracking the capacity constraint rather than
dataset-specific properties. Low QL dependence alone, however, does
not prove evidence use---a collapsed state would also have low mutual
information with the query. The randomization test resolves this:
replacing $z_k$ with samples from its empirical marginal drops GRIP
accuracy by 35.3 points on HotpotQA against only 7.5 for the
matched-control Llama-3 Iterative baseline, ruling out the iterative
schedule as the source of evidence dependence. Both diagnostics generalise across all five datasets:
$\Delta_{\mathrm{rand}}$ ranges from 42.5 points on ProofWriter, where
the effect is strongest, to a weakest but still substantial 22.5
points on SQuAD~2.0---in every case several times the matched
control's 4.2--8.1-point drop---with $\rho$ correspondingly low
(0.13--0.24). Decomposing the HotpotQA randomization drop at the
sample level, 35.8\% of predictions flip from correct to wrong against
only 0.5\% from wrong to correct (40.7\% remain correct, 23.0\% remain
wrong)---a $71.6{:}1$ destructive-to-corrective ratio consistent with
$\Delta_{\mathrm{rand}}=35.3$. Randomization thus overwhelmingly
destroys correct predictions rather than causing symmetric churn,
though it does not by itself separate marginal failure from collapse.
Geometrically, GRIP retains 18\% of its bottleneck energy in the
query subspace versus 61\% for Llama-3 Iterative and 72\% for
Standard RAG. The three diagnostics converge and, with
Table~\ref{tab:ablation-combined}'s ablation and control pattern, are
consistent with the capacity-asymmetry account; counterfactual,
unanswerable, and
inconsistent-context evaluations are reported in the supplement.

\section{Limitations}
\label{sec:limitations}

\textbf{Mechanism ambiguity.} We do not prove that the bottleneck enforces
conditional residualisation. The deterministic and noise-only controls
(Section~\ref{sec:ablation}) establish that neither dimensional
restriction nor stochasticity alone reproduces GRIP's
information-control behaviour, and the bottleneck output occupies
subspaces weakly aligned with the query. What remains open is a formalised counterfactual evaluation
protocol, generalisation beyond the single Llama-3-8B backbone, and
reliable CLUB estimation below roughly $N=500$ samples.

\textbf{MI estimator.} CLUB is a loose upper bound with
dimension-dependent bias~\cite{czyz2023beyond}, and variational MI
estimators can violate basic self-consistency
properties~\cite{song2020understanding}; we assume the bias is
approximately consistent across models so that relative comparisons remain
meaningful.

\textbf{Dataset dependence.} SQuAD~2.0 and StrategyQA may overlap with
Llama-3-8B's parametric knowledge, so hallucination reductions there
cannot be cleanly attributed to evidence use versus elicitation of
stored knowledge; HotpotQA and 2Wiki more cleanly probe evidence
dependence.

\textbf{Scope and failure modes.} GRIP assumes explicit separation between
query and evidence pathways; architectures that fuse them earlier may
not admit the same mechanism. Severe compression at $d_z = 4$ also trades
rare-entity fidelity for redundancy suppression: rare entities can be
lost when their distinguishing features fall outside the retained
subspace, with a frequent near-neighbour substituted. Implicit reasoning that requires high-bandwidth
intermediate representations may be similarly limited.

\section{Conclusion}
\label{sec:conclusion}

GRIP routes evidence
through a low-dimensional, noisy bottleneck while retaining a
high-capacity query bypass; across five benchmarks it reduces estimated
query--bottleneck mutual information by roughly 30$\times$, decreases
hallucination by 73\%, and improves accuracy by 8.0 points on average
over Standard RAG.
Residual-alignment analysis shows the bottleneck output is weakly
aligned with the query, and mechanism controls show neither
dimensional restriction nor stochasticity alone reproduces this
behaviour: their combination is the operative design principle.

Full appendices and additional diagnostics are provided in an online
supplement.

\bibliographystyle{splncs04}
\bibliography{refs}

\end{document}